\documentclass[10pt,twocolumn,letterpaper]{article}

\usepackage{cvpr} 

\usepackage[accsupp]{axessibility}  

\usepackage{graphicx}
\usepackage{microtype}
\usepackage[ruled,vlined]{algorithm2e}
\usepackage{algpseudocode}
\usepackage{soul}
\usepackage{colortbl}
\usepackage{bm}
\usepackage{comment}
\usepackage{amsfonts}
\usepackage{amsmath}
\usepackage{amssymb}
\usepackage{mathtools}
\usepackage{amsthm}
\usepackage{nicefrac}
\usepackage{bm}
\usepackage[misc]{ifsym}
\definecolor{Gray}{gray}{0.9}

\SetCommentSty{mycommfont}

\SetKwInput{KwInput}{Input}                
\SetKwInput{KwOutput}{Output}              

\definecolor{nblue}{rgb}{0.21,0.49,0.74}
\usepackage[pagebackref,breaklinks,colorlinks,allcolors=nblue]{hyperref}

\title{MotionMap: Representing Multimodality in\ Human Pose Forecasting}

\author{
\begin{tabular}[t]{c}
Reyhaneh Hosseininejad ${}^{* 1}$ \and Megh Shukla ${}^{* 1}$ \\
\end{tabular} \\
\begin{tabular}[t]{c}
Saeed Saadatnejad ${}^{1}$ \and Mathieu Salzmann ${}^{1, 2}$ \and Alexandre Alahi ${}^{1}$ \\
\end{tabular} \\
\\
${}^{1}$ École Polytechnique Fédérale de Lausanne (EPFL)\\
${}^{2}$ Swiss Data Science Centre (SDSC)\\
{\tt\small firstname.lastname@epfl.ch}
}

\newcommand\blfootnote[1]{%
  \begingroup
  \renewcommand\thefootnote{}\footnote{#1}%
  \addtocounter{footnote}{-1}%
  \endgroup
}

\begin{document}
\maketitle

\begin{abstract}
Human pose forecasting is inherently multimodal since multiple futures exist for an observed pose sequence. However, evaluating multimodality is challenging since the task is ill-posed. Therefore, we first propose an alternative paradigm to make the task well-posed. Next, while state-of-the-art methods predict multimodality, this requires oversampling a large volume of predictions. This raises key questions: (1) Can we capture multimodality by efficiently sampling fewer predictions? (2) Subsequently, which of the predicted futures is more likely for an observed pose sequence? We address these questions with MotionMap, a simple yet effective heatmap based representation for multimodality. We extend heatmaps to represent a spatial distribution over the space of all possible motions, where different local maxima correspond to different forecasts for a given observation. MotionMap can capture a variable number of modes per observation and provide confidence measures for different modes. Further, MotionMap allows us to introduce the
notion of uncertainty and controllability over the forecasted pose sequence. Finally, MotionMap captures rare modes that are non-trivial to evaluate yet critical for safety. We support our claims through multiple qualitative and quantitative experiments using popular 3D human pose datasets: Human3.6M and AMASS, highlighting the strengths and limitations of our proposed method. \\\url{https://vita-epfl.github.io/MotionMap} \blfootnote{${}^{*}$ Equal Contribution}

\end{abstract}

\section{Introduction}

\begin{figure}[!t]
  \centering
  \centerline{\includegraphics[width=\columnwidth]{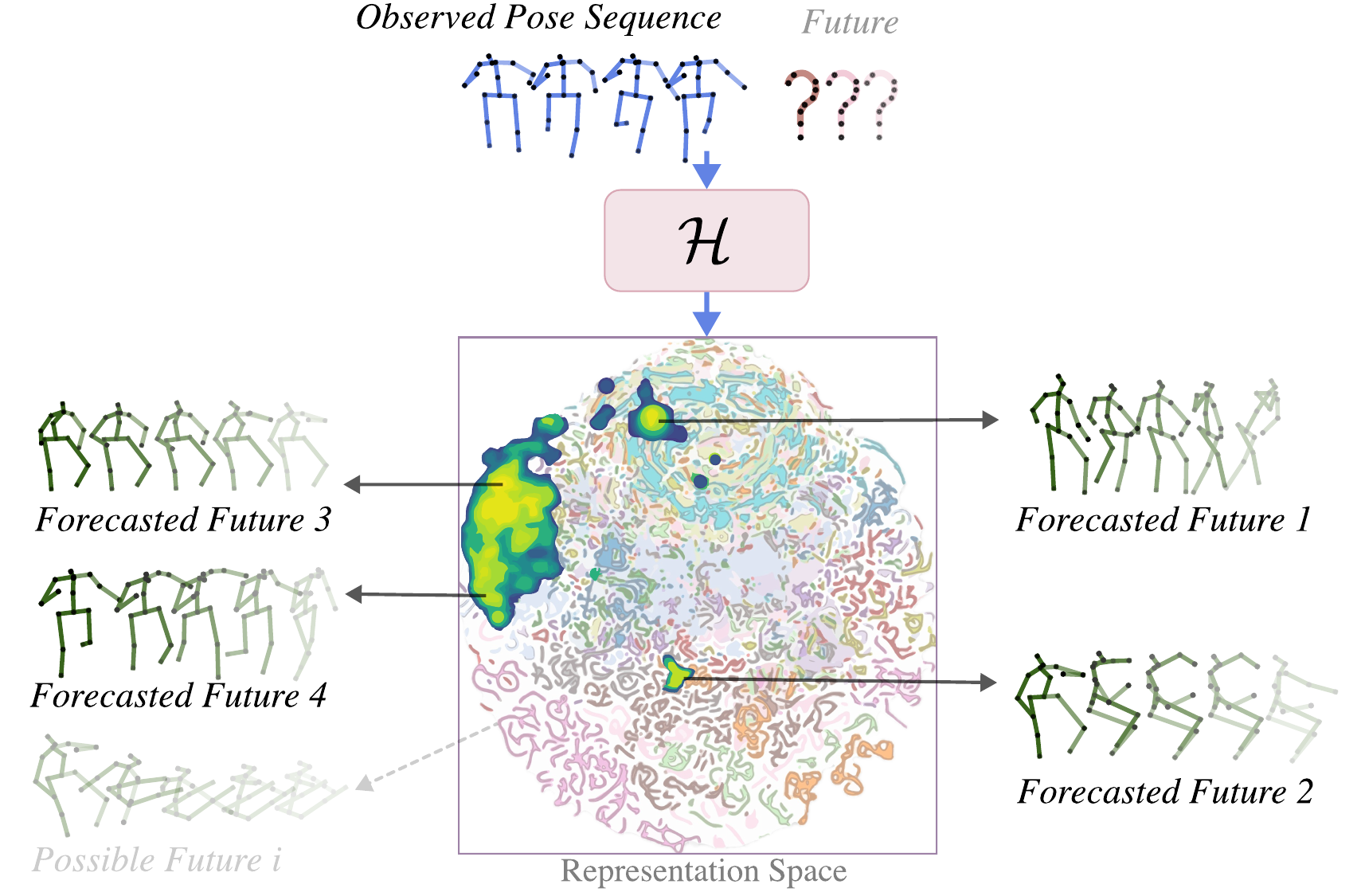}}
  \caption{\textbf{MotionMap} uses heatmaps to depict a spatial distribution over the space of motions. Local maxima imply that the corresponding motions have a higher likelihood of being a future motion for an observed pose sequence. MotionMap not only predicts a variable number of modes with the corresponding confidence, but it explicitly encodes rare modes that could otherwise be averaged out. 
  }
  \label{fig:motionmap}
\end{figure}

\begin{figure*}
\centerline{\includegraphics[width=0.8\paperwidth]{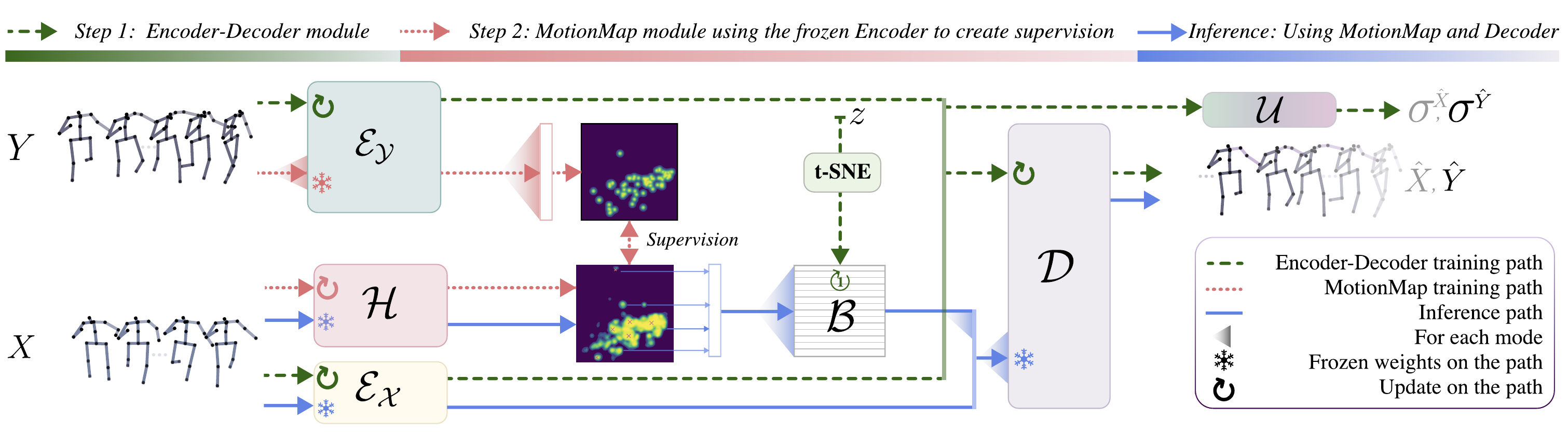}}
\caption{We define a two stage pipeline for human pose forecasting. At first, we train a framework similar to an autoencoder to predict the ground truth and future motion (Sec: 4.3). However, at test time we do not have the future motion and its latent as input. Therefore, we train a heatmap model to predict MotionMap, which along with the codebook encodes the likely motions and their latents as a drop-in replacement (Sec: 4.4). During fine-tuning and at inference time, we use the predicted MotionMap to obtain latents corresponding to motions with a high confidence and use it in tandem with the observed pose sequence to predict the future pose sequence (Sec: 4.5)}
\label{fig:flowchart}
\end{figure*}

Human pose forecasting is the task of predicting the future skeletal motion of a person given a set of past skeletal observations. The challenge arises from multimodality since an infinite number of futures with different levels of motion exist for the same observation. Typically, pose forecasting methods make a finite number of predictions to encompass these varied future motions. However, they can never cover all possible modes; one can trivially construct a set of ground-truth futures such that the model predictions have large errors. Indeed, this is reminiscent of the no-free-lunch theorem, which decries the existence of a universal learner. Therefore, one may wonder: Is human pose forecasting truly a solvable problem?

We therefore begin by proposing an alternate paradigm to make human pose forecasting well-posed. Instead of attempting to learn an unbounded set of future motions, we encourage the pose forecasting model to explicitly learn future motions present in the observed data. \textit{Specifically, models should use the training set to learn different transitions from input to output pose sequences and translate them to any unseen test sample.} By doing so the problem is no longer ill-posed, but well-posed since for every input sequence there exists a fixed number of futures, bounded by the size of the training set. Moreover, explicitly learning these transitions from the training dataset allows us to identify unknown motions at test time.

The question therefore is: How can we explicitly learn the different transitions within the observed data? The challenge is increased by the fact that the number of future motions is variable and dependent on the observed pose sequence. Moreover, state-of-the-art approaches~\cite{barquero2023belfusion, sun2024towards, yuan2020dlow, dang2022diverse} model multimodality by oversampling a large number of predictions. We therefore ask: (1) Can we capture multimodality by efficiently sampling a smaller number of predictions? (2) Subsequently, which of the predicted futures is more likely for an observed pose sequence?

In this paper, we propose \textbf{MotionMap}, a novel approach that captures different future \textit{motions} for each observation through a heat\textit{map}. Specifically, MotionMap interprets the heatmap as a spatial distribution over the space of all possible motion sequences in two dimensions. Different local maxima on this heatmap correspond to different possible future motions for an observed pose sequence (Figure \ref{fig:motionmap}). This representation has the primary advantage of encoding and predicting a variable number of modes for each observed sequence. Furthermore, we show that this representation allows us to explicitly learn different transitions from the training distribution and translate them for unseen test samples. MotionMap allows us to incorporate two different measures of confidence/uncertainty in our pose forecasting framework. These are the confidence of each mode as well as the uncertainty in the prediction conditioned on the mode. Finally, by design, MotionMap is sample efficient in achieving mode coverage for an observed pose sequence.

We perform experiments on two popular human pose forecasting datasets: Human3.6M and AMASS. Specifically, we compare different methods for (1) sample efficiency, where we compare the metrics for a fixed number of predictions; and (2) the ability to recall transitions present in the observed data. We make two observations: (1) MotionMap has the highest sample efficiency across all methods, and (2) MotionMap can accurately recall transitions from the observed data for unseen test samples. We make our code publicly available on our project page.

\section{Related Work}

\textbf{Architectures.} Early approaches employed feed-forward networks~\cite{li2018convolutional, guo2023back} and Recurrent Neural Networks (RNNs) to model the temporal aspects of the task~\cite{fragkiadaki2015recurrent, jain2016structural, martinez2017human}. Modeling temporal information implicitly is another effective method, commonly achieved by encoding each joint's trajectory with the Discrete Cosine Transformation (DCT), which helps mitigate common failures of auto-regressive models~\cite{mao2020history}. Subsequent advancements incorporated Graph Convolutional Networks (GCNs) to better capture the spatial dependencies of human poses~\cite{mao2019learning, cui2020learning, liu2021motion, sofianos2021space, ma2022progressively}. More recently, Transformers have shown effectiveness in capturing motion through serial spatial and temporal attention blocks~\cite{saadatnejad2024toward}, parallel spatial and temporal blocks~\cite{aksan2021spatio}, and hybrid structures~\cite{zhu2022motionbert}.

\textbf{Multimodality.} Recent pose forecasting methods acknowledge the inherent multimodality of human motion, aiming for a range of possible futures instead of a single outcome. This has been addressed through introducing stochasticity using generative models such as Generative Adversarial Networks (GANs)~\cite{barsoum2018hp, kundu2019bihmp}, Variational Auto-Encoders (VAEs)~\cite{walker2017pose, yan2018mt, cai2021unified, mao2021generating}, probabilistic latent variables~\cite{salzmann2022motron} and more recently, Diffusion models~\cite{chen2023humanmac, barquero2023belfusion, saadatnejad2023generic, sun2024towards}.  To promote diversity across the samples, DLow~\cite{yuan2020dlow} introduced a new sampling strategy on top of conditional VAEs by generating multiple Gaussian distributions and then sampling latent codes from different Gaussian priors. GSPS~\cite{mao2021generating} presented a two-stage sampling strategy, addressing the lower and upper body separately. DivSamp~\cite{dang2022diverse} proposed using the Gumbel-Softmax sampling strategy from an auxiliary space, and STARS~\cite{xu2022diverse} introduced an anchor-based sampling method. Other methods such as SLD~\cite{xu2024learning} projects motion queries into a latent space to allow for more diverse motions, and MDN~\cite{kim2024mdn} proposed a transformer-based mechanism that enhances sampling diversity in the latent space. To generate more realistic forecasts, TCD~\cite{saadatnejad2023generic} introduced a temporally-cascaded diffusion model that handles both perfect and imperfect observations, and BelFusion~\cite{barquero2023belfusion} presented a conditional latent diffusion model and recently, CoMusion~\cite{sun2024towards} introduced a single-stage stochastic diffusion-based model using both Transformer and GCN architectures.

\textbf{Limitations.} Although these methods achieve notable results, there exists certain limitations. First, diversity does not entail realism. While diversity based methods score high on diversity, the forecasts often lack coherence with the observation. Second, stochastic approaches commonly learn an \textit{implicit} distribution over the multimodal future. As a result, they require large-scale sampling for a wider mode coverage. Moreover, the number of samples to draw for different observations is unknown. Further, are all samples equal, or are some more likely than the other? We argue for the need to have better representations for multimodality to address these limitations. However, we first begin by asking, what is multimodality in human pose forecasting?

\section{Multimodality in Human Pose Forecasting}

Without a concrete definition, we interpret a mode as a \textit{set of motions corresponding to an action which are likely for a given observation}. For instance, walking represents a mode with various walking styles as samples. We add \textit{likely} to avoid ambiguities, for example, related to left and right-handed actions within a mode. If the observation is left-handed, it is unlikely that the forecast should be right-handed. Therefore,

\begin{quote}
    \centering
    \textit{Multimodality in human pose forecasting refers to a diverse yet realistic set of future actions with a logical transition from an observed pose sequence.}
\end{quote}

Subsequently, we then wonder: How do we obtain multimodal ground truths for an observed pose sequence? This is non-trivial since each input pose sequence is paired with a single output sequence, which is the actual future motion. Therefore, recent literature \cite{barquero2023belfusion} computes multimodal ground truth by comparing a given input pose sequence with other input pose sequences in the dataset. If two input pose sequences are similar, the corresponding future motions of each sequence can be considered as the multimodal ground truth for each other. This similarity is measured based on a threshold on the distance between the last frame of the two observations.

\begin{figure}[!t]
\centerline{\includegraphics[width=0.65\columnwidth]{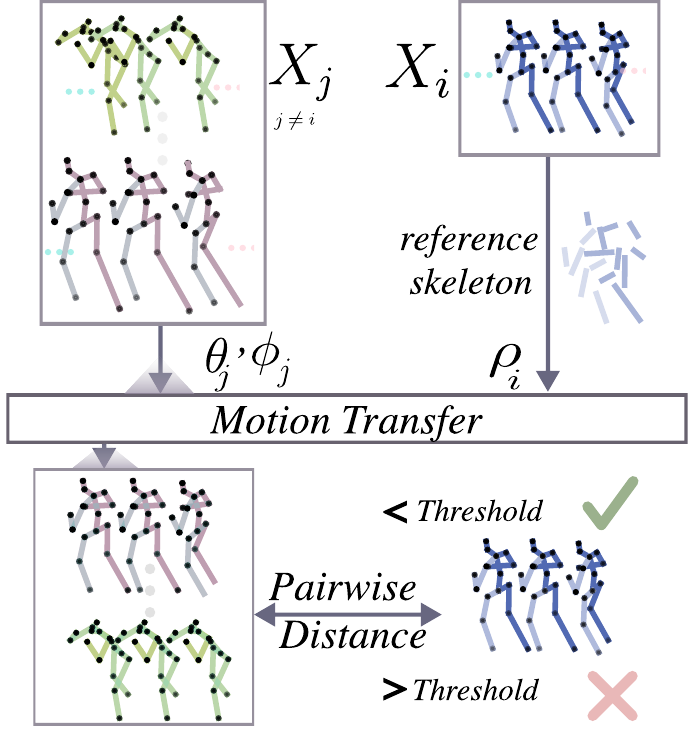}}
\caption{The current approach to finding multimodal ground truths uses only the last frame to measure the similarity between sequences. However, not only does this lose out on motion information, but persons of different sizes with the same motion may not be considered for multimodal ground truth. Hence, we propose computing the ground truths by using the last three frames and scaling the skeleton while retaining the motion. We do this using cartesian to spherical coordinate transformations.}
\label{fig:mmgt}
\end{figure}

\begin{figure*}
\centerline{\includegraphics[width=1.8\columnwidth]{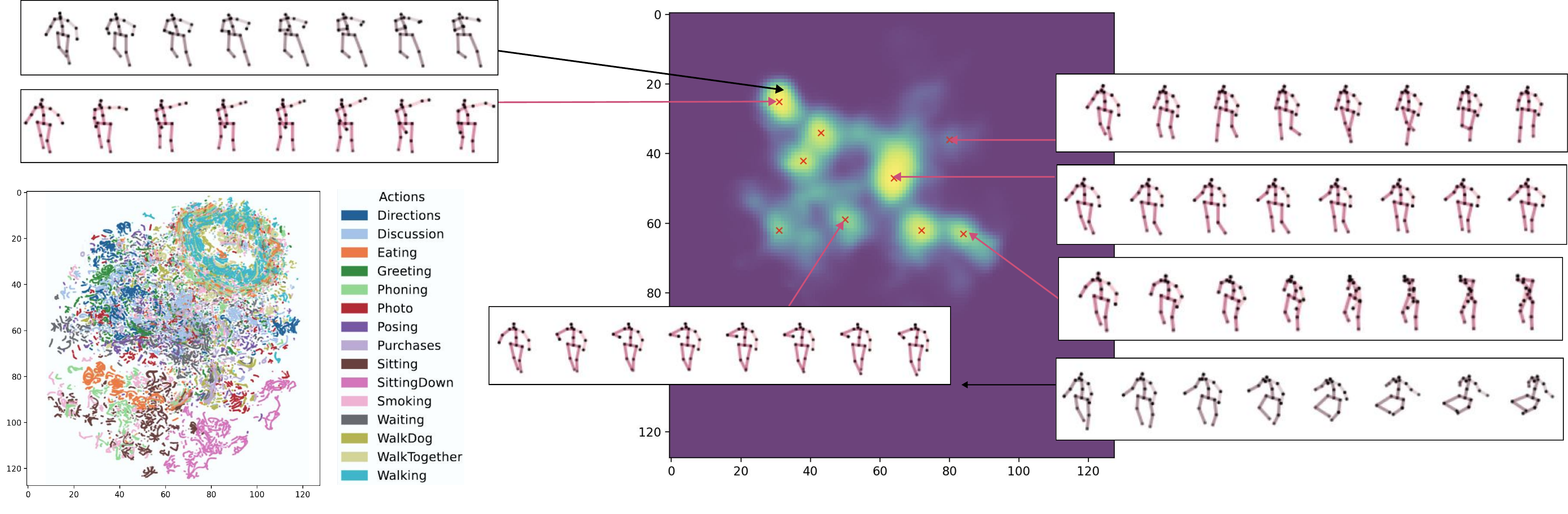}}
\caption{\textbf{Controllability.} 
MotionMap can also be used with auxiliary data such as action labels for controllable pose forecasting. Since each pose sequence is associated with an embedding and action label, a spatial distribution over the space of motions is the same as that over the action labels. This allows for the use of MotionMap to select modes based on the confidence as well as user preference for the forecasted action. We illustrate this distribution over the space of motions $\leftrightarrow$ actions for an example input from the Human3.6M dataset.}
\label{fig:control}
\end{figure*}

However, this approach has two main limitations. First, it does not normalize skeletal sizes across different individuals. Consequently, pose sequences with similar motions but different skeletal sizes are not recognized as multimodal ground truths for each other. Second, relying only on the final input frame can lead to abrupt transitions in multimodal ground truths, as a single frame contains positional but not motion-related information. To address these issues, we modify the multimodal ground truth definition by (1) incorporating the last three frames (2) scaling skeletons of other pose sequences to match the given sequence (Figure \ref{fig:mmgt}).

Skeleton scaling is achieved by converting poses from Cartesian to spherical coordinates. Specifically, we represent each joint (e.g., elbow) in spherical coordinates relative to its parent joint (e.g., shoulder), preserving angular positioning while allowing adjustments in joint length. By substituting the original joint lengths with those from the reference skeleton, we scale the entire pose sequence accordingly. Pseudo-code detailing this scaling procedure is provided in Appendix \ref{app:motion}.

\section{Methodology}

Using our definition of the multimodal ground truth, let us discuss how we encode them through MotionMap.

\textbf{Notation.}  Let $X$ be the sequence of observed poses consisting of $T_o$ frames
$X = [x_1,\dots,x_{T_o}]$. Each pose $x_i$ is of dimension $(J,3)$ where $J$ is the number of joints. Similarly, $Y$ is defined as the sequence of future poses with $T_f$ frames, $Y=[x_{T_o+1},\dots,x_{T_o+T_f}]$. We define the set of multimodal ground truths as $\bm{{Y}_{mm}}: \{{Y^i}|i:1,.., M\}$,  where ${M}$ is the number of ground-truth modes. We note that the number of ground-truth modes varies for different samples. During inference, the pose forecaster receives $X$ as input and generates ${\bm{\hat{Y}_{mm}}}: \{\hat{Y^i}|i:1,.., \widehat{M}\}$, where $\widehat{M}$ is the predicted number of modes. We also note that ${M}$ and $\widehat{M}$ could also differ.

\textbf{Overview.} Our pose forecasting framework, depicted in \Cref{fig:flowchart}, consists of two trainable modules. We have an autoencoder consisting of Gated Recurrent Unit encoders $\mathcal{E}_{\mathcal{X}}$, $\mathcal{E}_{\mathcal{Y}}$, a simple multilayer perceptron based uncertainty estimation module $\mathcal{U}$, and a GRUCell based decoder $\mathcal{D}$. We use their architectures from \cite{barquero2023belfusion}, and provide a detailed description in appendix \ref{app:implementation}. The autoencoder is not only used for pose forecasting but also to obtain intermediate representations of different output pose sequences, which, as we shall see later, are used to learn the MotionMap. Our second trainable module is $\mathcal{H}$, which is used to predict the MotionMap for a given observation. Our training process therefore consists of two steps to learn the two modules. 

\subsection{Step 1: Autoencoder Module}

The two encoders $\mathcal{E}_X$ and $\mathcal{E}_Y$ take as inputs $X$ and $Y$ respectively. If $X$ has multiple multimodal ground truths $\bm{{Y}_{mm}}$, then we select one $Y^i$ randomly, with different ground truths selected across different epochs. This is equivalent to a Monte-Carlo approximation of the multimodal distribution for each sample. The outputs of these two networks are arrays $z_x$ and $z_y$ which are concatenated and passed through a simple two layer multilayer perceptron non-linearity giving us $f(z_x \bigoplus z_y)$. Subsequently, this new array is passed to the decoder $\mathcal{D}$ to predict the entire sequence $\widehat{X \bigoplus Y_i}$. The reasoning behind predicting a concatenation of the input and multimodal ground-truth sequence is to avoid discontinuities and predict a smoother transition from the input pose sequence to the predicted pose sequence. This prediction along with the uncertainty predictions from $\mathcal{U}$ are optimized using the negative log-likelihood.

\textbf{Uncertainty.} We use $f(z_x \bigoplus z_y)$, the input to the decoder, to also predict the corresponding uncertainty of the prediction $\bm{\sigma}^2 \in \mathbb{R}^{(T_o + T_f) \times J}$. We use a normal distribution to model the uncertainty where both the mean and variance are predicted. If we define the mean square error \textit{per joint} as $\text{error} = || \widehat{X \bigoplus Y_i} - X \bigoplus Y_i ||_2^2$, the training objective is
\begin{equation}
\mathcal{L} = \dfrac{\text{error}}{\bm{\sigma}^2} + \log \bm{\sigma}^2.
\label{eq:nd}
\end{equation}
We minimize this loss to jointly optimize $\mathcal{U}$ and $\mathcal{D}$, completing the training step for the autoencoder. While we follow ~\cite{stirn2023faithful} for simplicity, advanced techniques for heteroscedastic regression ~\cite{seitzer2022on, immer2023effective, shukla2024tictac, shukla2025towards} are recommended.

\begin{figure*}[!ht]
\centerline{\includegraphics[width=0.63\paperwidth]{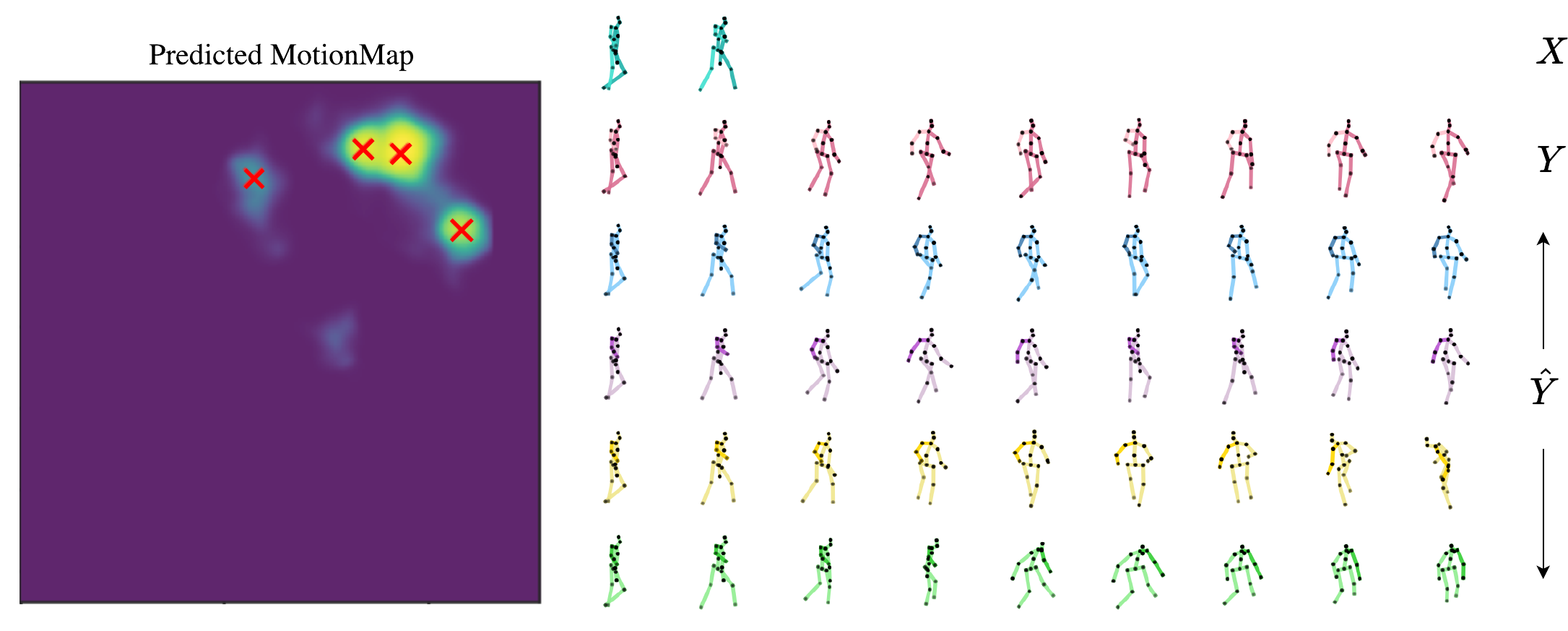}}
\caption{\textbf{Ranking}. Since MotionMap can predict variable number of modes with their associated confidences, our method also allows us to rank predictions. For instance, the highest ranked prediction (top row among $\hat{Y}$) closely matches the ground truth motion. However, rare modes (bottom row) are ranked low since the corresponding mode has lower confidence.}
\label{fig:ranking}
\end{figure*}

\subsection{Step 2: MotionMap Module}

If we were to use the autoencoder as a pose forecaster at test time, we would be required to know the joint embedding $z_x \bigoplus z_y$, which is the input to the decoder. However, although we can compute $z_x$ since we know the input pose sequence, $z_y$ is nontrivial to obtain. While the literature utilizes various approaches stemming from generative models to obtain this latent, we propose MotionMap, which uses heatmaps and codebooks to represent various motion sequences. \textit{Intuitively, MotionMap can be interpreted as a learnt prior over the future motions for each sample, identifying the spread of modes and their associated likelihood.} Given $\bm{\hat{Y}_{mm}}$, we encode each multimodal ground truth on a two-dimensional heatmap, allowing us to represent a variable number of futures per sample. However, how can we trace these future pose sequences to two dimensional coordinates and then back to $z_y$?

\subsubsection{Dimensionality Reduction}

\label{sec:dimred}

We obtain two dimensional embeddings by first encoding all futures $Y$ across the dataset using $\mathcal{E}_y$ to $z_y$. Next, we project all embeddings $z_y$ using t-SNE ~\cite{vandermaaten2008tsne} into two dimensions. We choose t-SNE because it is a popular non-linear technique with an optimized implementation \cite{openTSNE}.  Subsequently, we scale the two dimensional embeddings such that they span the size of the heatmap. Finally, we quantize the embeddings by rounding them to the nearest integer to obtain $h_y$, of dimensionality two.

To create the heatmap, we iterate over each of the $M$ futures in $\bm{\hat{Y}_{mm}}$, reducing them to their corresponding $h_y$, and plot a Gaussian centered around it. Such an approach ensures that variable number of modes $M$ can be plotted per sample. Additionally, very similar futures are mapped to the same $h_y$, avoiding duplicity of ground truths. As a result, rare modes are not suppressed in MotionMap.

To go back to $z_y$, we create a mapping between the embedding $h_y$ and $z_y$ which is stored in a codebook $\mathcal{B}$.  Since dimensionality reduction maps from the large volume source domain to a smaller volume target domain, there exist multiple $z_y$ that are mapped to the same $h_y$. Therefore, when creating the codebook, we map every $h_y$ to the \textit{mean} of all $z_y$ that are mapped to it, giving us $\mathcal{B} = \{h_y: \overline{z_y}\}$. The size of the codebook is independent of the dataset and depends on the level of quantization of the heatmap and the size of the embedding.  For dimensionality $m \times m$, and using 32-bit floating point embeddings of length $n$, the size corresponds to $\approx 32m^2n$ bits. For $m=n=128$, the size is 64MB.

\begin{figure*}
\centerline{\includegraphics[width=1.6\columnwidth]{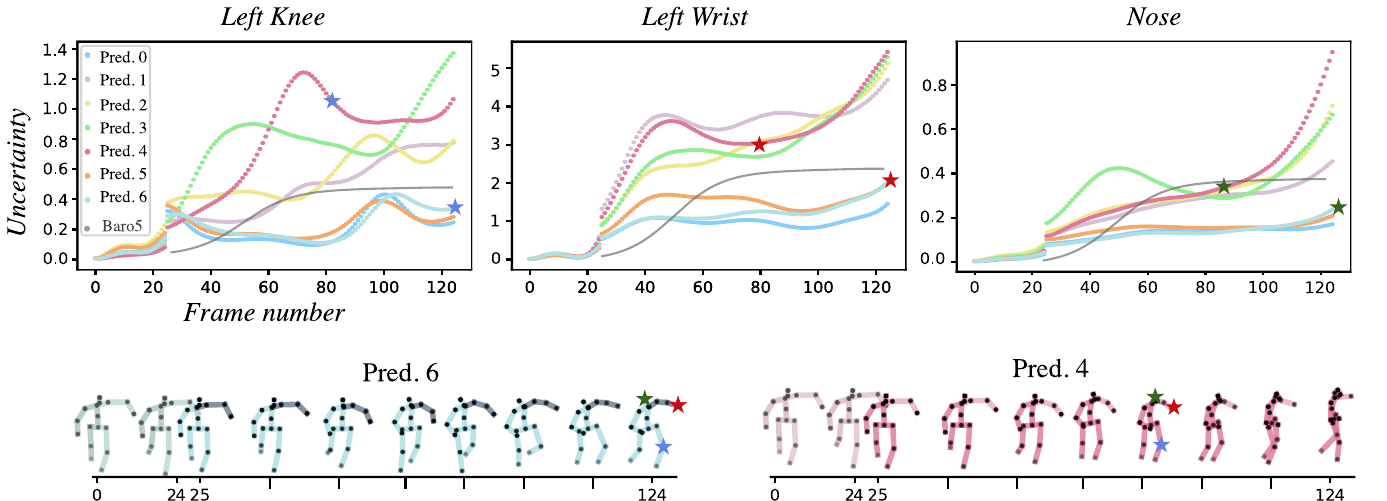}}
\caption{\textbf{Uncertainty.} Previous work \cite{saadatnejad2024toward} assumed homoscedasticity, learning generic uncertainty trends independent of the input. In contrast, we show that heteroscedastic modeling can result in semantically richer uncertainty estimates (Sec 5.3). We do this by decomposing uncertainty into that of the mode (through MotionMap) and the forecast conditioned on the mode (through the uncertainty module $\mathcal{U}$). For instance, prediction 6 has a lower uncertainty in comparison to prediction 4 because the latter involves fast changes in direction.}

\label{fig:u0}
\end{figure*}

\subsubsection{Training}

Training the heatmap predictor $\mathcal{H}$ to obtain MotionMaps for different samples is depicted in \Cref{fig:flowchart}. For a given sample $X$, the MotionMap model $\mathcal{H}$ is trained to minimize the pixel-wise binary cross-entropy loss with the heatmap constructed using ground-truth multimodal samples. To prevent overfitting, the MotionMap model uses a simple GRU encoder which spatio-temporally encodes the last three frames of the observation $X$. This is followed by a single fully connected layer, which culminates in a series of 1x1 convolutions. The architecture is detailed in the appendix.

\subsection{Fine-Tuning} 
To improve the performance at test time, we recommend fine-tuning $f(\cdot)$ and $\mathcal{D}$. This is because the codebook maps the heatmap to the \textit{average} embedding, $\mathcal{B} = \{h_y: \overline{z_y}\}$. The fine-tuning is similar to training, with the exception that we use $\overline{z_y}$ instead of $z_y$. This requires mapping a multimodal future to its heatmap embedding, and using the codebook to decode the average latent.

\subsection{Inference}

At test time, we use the predicted MotionMap coupled with the codebook to obtain the missing latent $\overline{z_y}$. Specifically, given an observation $X$, we predict the corresponding MotionMap using the model $\mathcal{H}$. We then compute various local maxima \textit{deterministically} which correspond to the likeliest modes of the future sequences. The number of maxima $\widehat{M}$ can not only vary for different observed sequences but may also differ from the number of ground truth modes $M$. This is because we use weighted binary cross-entropy to penalize false negatives more than false positives. This leads to potential new modes that emerge across the dataset.

Next, for each mode, we use the codebook $B$ to index the missing latent $\overline{z_y}$. We also obtain $z_x$ from $\mathcal{E}_X$, allowing us to pass the updated latent $f(z_x \bigoplus \overline{z_y})$ to the decoder for pose forecasting. Finally, we note that this methodology does not depend on stochasticity.

\section{Experiments} \label{sec:imp_det}

\textbf{Overview.} We study\footnote{\url{https://www.github.com/vita-epfl/MotionMap}} human pose forecasting as a well posed problem, which is to learn and translate different unseen transitions from the observed dataset to unseen test samples. We perform extensive quantitative and qualitative analysis related to controllability, ranking, and uncertainty estimation. We use the AMASS ~\cite{mahmood2019amass} and the Human3.6M ~\cite{ionescu2013human3} datasets. We describe these implementation and dataset in detail in Appendix \ref{app:implementation} and \ref{app:dataset}. Our evaluation uses metrics from prior work ~\cite{yuan2020dlow}. Average Displacement Error (ADE) measures the minimum L2 distance between ground truth and predicted sequences averaged over all frames, while Final Displacement Error (FDE) considers only the last frame. This paper specifically emphasizes their multimodal counterparts, MMADE and MMFDE, which average ADE and FDE across multiple ground truths. We apply standard thresholds of 0.5 for Human3.6M and 0.4 for AMASS multimodal ground truth selection. 

\begin{figure*}[t!]
\centerline{\includegraphics[width=2.\columnwidth]{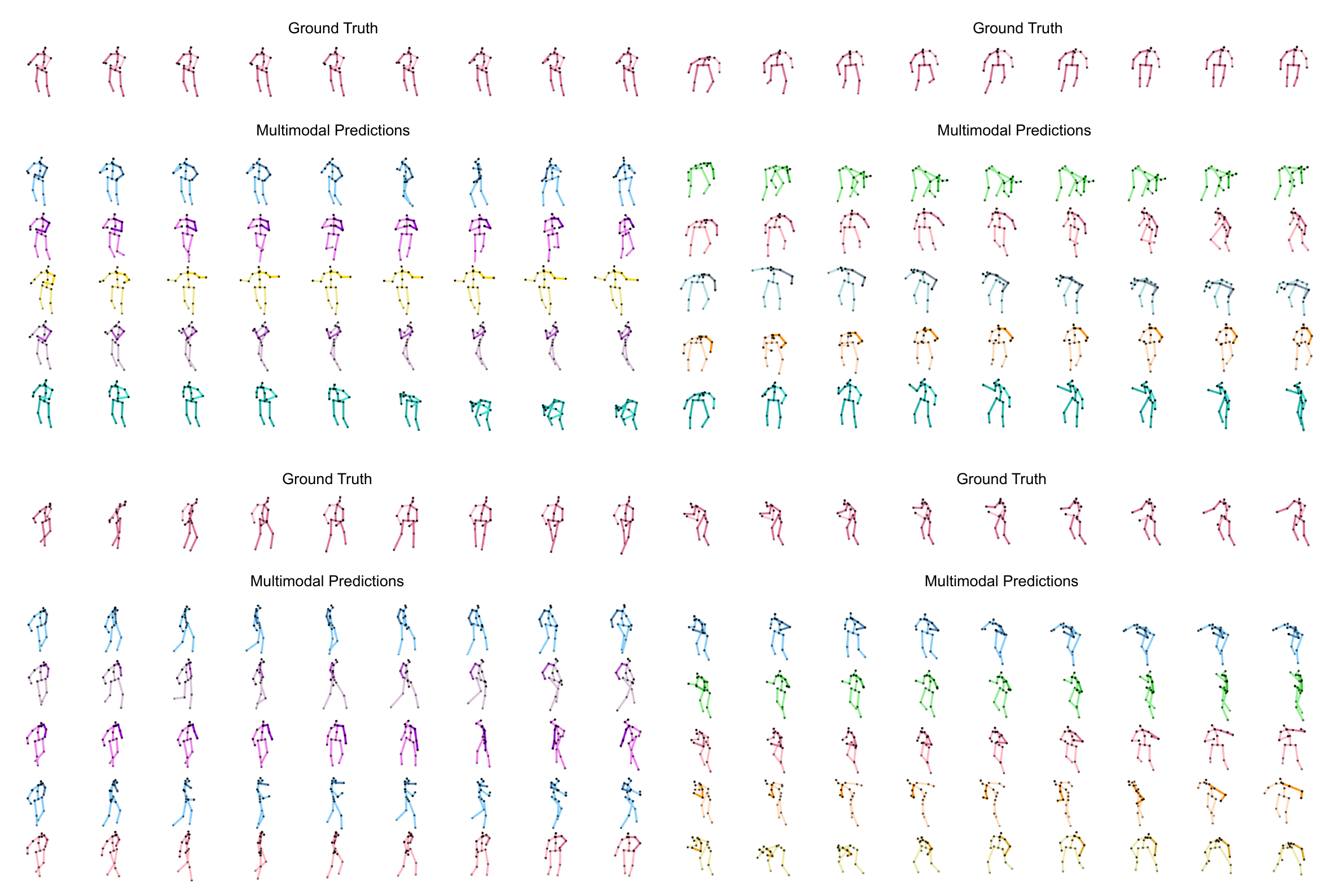}}
\caption{\textbf{Diversity}. By virtue of using the same decoder as BeLFusion \cite{barquero2023belfusion}, our method learns realistic yet diverse motions. This is because MotionMap can be decoded to select modes that are sufficiently different yet likely.}
\label{fig:q}
\end{figure*}

\subsection{Controllability}

Figure \ref{fig:control} demonstrates that pose forecast predictions can be done based on user preferences. For instance, metadata such as action labels can be used to select the most likely pose forecasts that align with user intent. These preferences may stem from desired actions extracted via language models to generate specific pose forecasts. Additionally, forecasts similar to a selected one can be obtained by choosing the latent embedding corresponding to the nearest neighbors of a local maximum. Additional figures are in Appendix \ref{app:controllability}

\subsection{A Tale Of Two Uncertainties}

In our approach, we learn both the uncertainty of each prediction and the confidence associated with each mode. Specifically, we model the relationship as $X \rightarrow Z \rightarrow Y$, where $Z$ represents the mode and $Y$ represents the forecast. Introducing $Z$ simplifies a multimodal distribution over possible futures into an approximately unimodal distribution. This improves upon homoscedastic modeling, which treats uncertainty as independent of the observed pose \cite{saadatnejad2024toward} and averages uncertainty across all future motions, leading to generic trends such as increasing uncertainty over time.

In contrast, MotionMap captures both action ambiguity and motion-specific uncertainty. It models a spatial distribution over possible actions that also gives the confidence per mode. We also condition on specific modes to limit future motions, resulting in uncertainty trends tailored to the prediction. For instance, the nose joint primarily provides skeleton orientation. In Figure \ref{fig:u0}, \textit{Pred. 4} exhibits greater nose uncertainty than \textit{Pred. 6}, due to a sharper turn. Additionally, increased uncertainty during the transition (frames 25-30) prevents motion discontinuities between the observed sequence and different multimodal ground truths across various subjects. We also note that joints with higher mobility generally display greater uncertainty. Additional visual examples are provided in Appendix \ref{app:uncertainty}.

\subsection{Sampling Comparison}

\begin{figure*}
\centerline{\includegraphics[width=\linewidth]{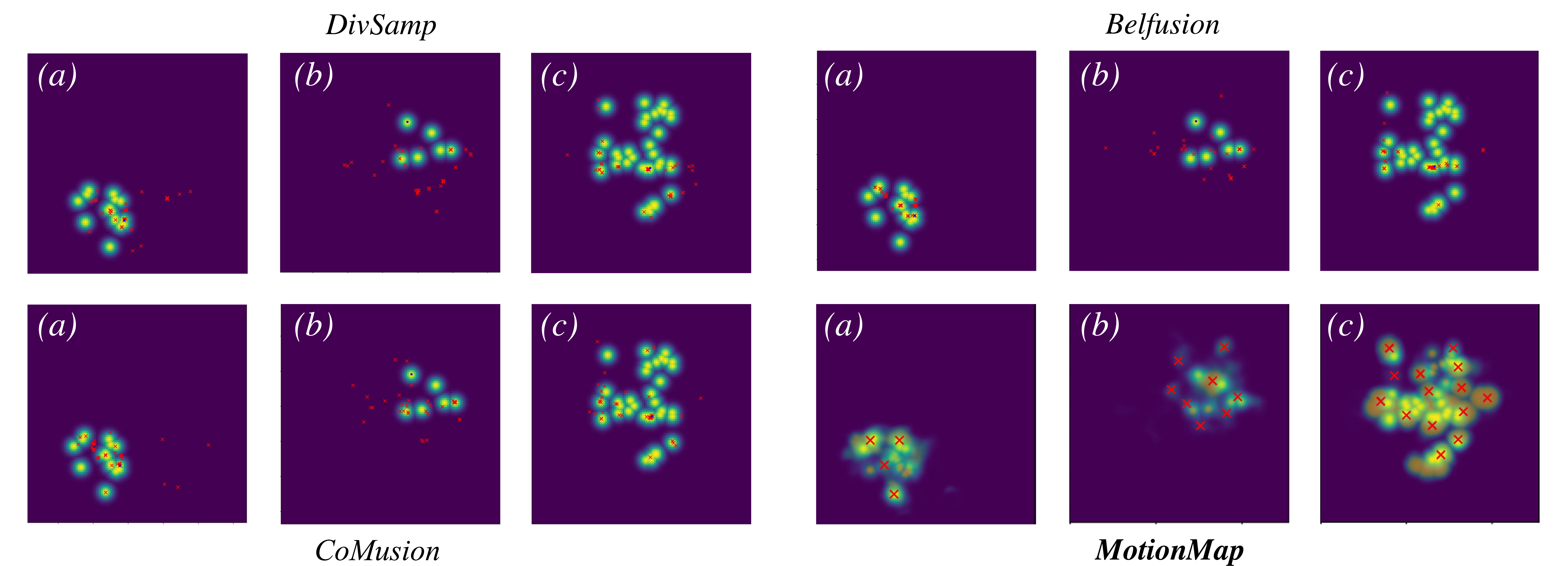}}
\caption{\textbf{Sampling comparison.} We compare the pose forecasts of different methods using the MotionMap representation for three different observations \textit{(a, b, c)}. The multimodal ground truth is represented through MotionMap, and the red points are two-dimensional projections of the predicted futures. For the MotionMap method, we overlay the predicted MotionMap itself. We observe that the MotionMap model performs better since the predicted MotionMap aligns with the ground truth. \textit{Please zoom!}}

\label{fig:sampling_comparison}
\end{figure*}

Can we use the MotionMap representation to compare multimodality across state-of-the-art? To do so, we collected predictions for each baseline for three different sequences. We then encode these predictions into two dimensions as described in Section \ref{sec:dimred}. Next, we overlay them on the ground truth MotionMap to identify the differences in the predictions and the ground truth. We observe in Figure \ref{fig:sampling_comparison} that baselines that rely on anchors, although diverse, predict transitions that are unlikely for the given pose sequence. Diffusion based methods are less diverse and do not capture rare modes. In contrast, MotionMap predictions captures the spread of the multimodal ground truth. The full comparison is in Appendix \ref{app:sampling}.

\subsection{Ranking Predictions and Diversity}

Unlike state-of-the-art methods, MotionMap predicts the confidence of different modes and allows us to rank our predictions in order of confidence. In Figure \ref{fig:ranking}, we plot the predicted MotionMap along with the associated pose forecasts, ranked based on the confidence of the mode. We observed that high confidence modes are often correlated with the ground truth future, and low confidence modes indicating rare transitions from the observed pose. We also visualize some examples of the generated poses in \Cref{fig:q}. Given input sequences, we see that our generated pose sequences are markedly diverse while also depicting smooth and realistic motions.

\begin{table*}[!t]
  \caption{\textit{Human3.6M dataset and AMASS dataset.} All baselines (except for zero-velocity) are limited to 7 forecasts. Our method, unconstrained by the number of modes, is adjusted to produce an equal number of predictions. Metrics are reported in meters.}
  \label{table:q2}
  \centering
  \small
  \resizebox{\textwidth}{!}{
  \begin{tabular}{l!{\color{gray}\vline}c c c c c!{\color{gray}\vline}c c c c c}
    \toprule
    \multicolumn{1}{c!{\color{gray}\vline}}{} & \multicolumn{5}{c!{\color{gray}\vline}}{\textbf{Human3.6M}} & \multicolumn{5}{c}{\textbf{AMASS}} \\
    \textbf{Method} & Diversity $\uparrow$ & ADE $\downarrow$ & FDE $\downarrow$ & MMADE $\downarrow$ & MMFDE $\downarrow$ & Diversity $\uparrow$ & ADE $\downarrow$ & FDE $\downarrow$ & MMADE $\downarrow$ & MMFDE $\downarrow$ \\[2pt]
    
    \toprule
    \rowcolor{Gray} Zero-Velocity & - & 0.597 & 0.884 & 0.617 & 0.879 & - & 0.755 & 0.992 & 0.778 & 0.996 \\
    TPK~\cite{walker2017theposeknows} & 6.53 & 0.534 & 0.691 & 0.559 & 0.675 & 8.57 & 0.519 & 0.634 & 0.600 & 0.678 \\
    \rowcolor{Gray} DLow~\cite{yuan2020dlow} & 11.77 & 0.445 & 0.730 & 0.576 & 0.715 & 12.69 & 0.471 & 0.594 & 0.554 & 0.633 \\
    GSPS~\cite{mao2021gsps} & 14.97 & 0.512 & 0.684 & 0.550 & 0.665 & 13.55 & 0.501 & 0.662 & 0.591 & 0.688 \\
    \rowcolor{Gray} DivSamp~\cite{dang2022diverse} & \textbf{15.73} & 0.480 & 0.685 & 0.542 & 0.671 & \textbf{25.90} & 0.479 & 0.638 & 0.623 & 0.728 \\
    STARS~\cite{xu2022diverse} & 15.85 & 0.508 & 0.697 & 0.551 & 0.686 & - & - & - & - & - \\
    \rowcolor{Gray} TCD~\cite{saadatnejad2023generic} & 7.63 & 0.603 & 0.828 & 0.668 & 0.835 & - & - & - & - & - \\
    BeLFusion~\cite{barquero2023belfusion} & 7.11 & 0.441 & \textbf{0.597} & 0.491 & 0.586 & 7.92 & 0.347 & 0.478 & 0.488 & 0.564 \\
    \rowcolor{Gray} CoMusion~\cite{sun2024towards} & 7.32 & \textbf{0.426} & 0.613 & 0.531 & 0.623 & 7.39 & \textbf{0.311} & 0.460 & 0.526 & 0.602 \\
    \midrule
    \textbf{MotionMap} & 7.84 & 0.474 & \textbf{0.598} & \textbf{0.466} & \textbf{0.532} & 8.50 & 0.325 & \textbf{0.450} & \textbf{0.450} & \textbf{0.514} \\
    \bottomrule
  \end{tabular}
  }
\end{table*}

\subsection{Quantitative Results}

We quantitatively assess different baselines on their ability to translate multimodality from observed data to any \textit{unseen test sample}. To achieve this, we identify multimodal ground truths in the training labels that are closest to each test sample. This approach not only ensures a well-posed problem but also mitigates discrepancies between training and testing distributions, as demonstrated in the appendix (Figure \ref{fig:hmyo2}). Moreover, in such cases, the limited size of the test set may fail to capture the full diversity of modes present in the training data. For instance, if the test split contains only five samples, each test sample will have at most five and at least one multimodal ground truth, making the evaluation inherently unreliable.

We compare various baselines in \Cref{table:q2} using the Human3.6M and AMASS datasets. While methods such as DLow and DivSamp are diverse, they do not accurately predict the futures corresponding to the observed motion, since not all anchors in the latent are equally likely. These methods also tend to be more diverse than the ground truth itself. In contrast, by imposing a `prior' on the latent, MotionMap successfully predicts the likeliest modes while being sample efficient. We additionally note that MotionMap successfully outperforms strong baselines in extending multimodal transitions in the observed data to all test samples. We also highlight a comparison of MotionMap with BeLFusion \cite{barquero2023belfusion}. Since the two share the same encoder and decoder, the major difference is in generating $z_y$. BeLFusion uses diffusion and repeated sampling whereas MotionMap uses heatmaps and codebooks, providing significant gains.

Finally, we also report the performance using the existing methodology of restricting the multimodal ground truth to the testing split (Tables \ref{table:app_h36m}, \ref{table:app_amass}). We observe that while MotionMap is more accurate in recalling transitions from the test set, this does not come at the cost of general performance for unseen motions.

\textbf{Limitations.} A primary limitation of our method is the lack of fine-grained motion prediction within a mode, such as predicting subtle variations in walking. We particularly observe this effect in the testing split with many of the multimodal ground truths being time-shifted versions of the observation. These ground truths are clumped by MotionMap under one mode. Since MotionMap encodes only one forecast per mode, this may result in higher ``errors" since the forecast cannot explain all subtle variations with the samples. However, the choice of clumping similar motions together remains a design choice of MotionMap.

\section{Conclusion}
In this work, we discussed a paradigm to make human pose forecasting well-posed. Next, we proposed a new representation that explicitly encodes a varying number of future motions for different observations. The representation also quantifies the confidence of different modes, allowing us to identify more likely futures from less likely ones. With MotionMap, we learned diverse yet realistic motion transitions from the observed data. By explicitly predicting the spread of future motions, MotionMap is sample-efficient and does not rely on repeated random sampling to achieve mode coverage. We highlighted a potential use of MotionMap for controllable pose forecasting, as well as uncertainty estimation from a practical standpoint. Through qualitative and quantitative analysis, we highlighted the effectiveness of MotionMap for robust and safe human pose forecasting.

\subsection*{Acknowledgment}

This research is funded by the Swiss National Science Foundation (SNSF) through the project \textit{Narratives from the Long Tail: Transforming Access to Audiovisual Archives} (Grant: CRSII5\_198632). The project description is available on:\\\url{https://www.futurecinema.live/project/}

\bibliographystyle{IEEEtran}
\bibliography{arxiv}

\newpage
\appendix
\pagenumbering{gobble}
\section*{Appendix}

\textbf{We make the code available on \url{https://github.com/vita-epfl/MotionMap}.}
\section{Algorithm: Motion Transfer}
\label{app:motion}

We use `motion transfer' to ensure that different labels $y$ have the same skeletal size as that of the input $x$. This is done to ensure that multimodal ground truths are selected purely on the motion, and not on the size of the person. We also follow this reasoning and use motion transfer during the training and evaluation process.

\begin{algorithm}
\small
\tcc{Transfers the motion from pose sequence $y$ to skeleton of sequence $x$}
\vspace{0.2cm}
\tcp{Pose sequence for skeletal reference}
\KwInput{$x \in \mathbb{R}^{\#\text{frames}_x \times \#\text{joints} \times 3}$}
\tcp{Pose sequence for motion reference}
\KwInput{$y \in \mathbb{R}^{\#\text{frames}_y \times \#\text{joints} \times 3}$}

\tcp{Pose sequence with skeletal size from $x$ and motion from $y$}
\KwOutput{$z \in \mathbb{R}^{\#\text{frames}_y \times \#\text{joints} \times 3}$}
\vspace{0.2cm}

\tcp{Get last frame}
pose = x[-1] \\
\tcp{Get length for each link in pose}
$\rho$, $\_$, $\_$ = \texttt{cartesian\_to\_spherical}(pose) \\
\tcp{Get motion (angles) for each link in pose}
$\_$, $\theta$, $\phi$ = \texttt{cartesian\_to\_spherical}(y) \\
\tcp{Reconstruct new pose sequence}
$z$ = \texttt{spherical\_to\_coordinates}($\rho$, $\theta$, $\phi$)

\Return $z$

\caption{\textbf{\textit{Motion Transfer}}} \label{algo:mt}
\end{algorithm}

\section{Implementation Details}

\label{app:implementation}
We base most of our architecture on those proposed in \cite{barquero2023belfusion}. Our encoders $\mathcal{E}_X$ and $\mathcal{E}_Y$ are based on gated recurrent units (GRU) with a dimensionality of 128. Our pose forecaster $\mathcal{D}$ is the exact same design as BeLFusion. The major difference is that we predict a concatenation of the input and output sequence. The uncertainty module is a simple multilayer perceptron (MLP) that predicts the uncertainty per joint per time frame. The heatmap model uses a combination of the GRU encoder, and a one layer MLP, and gives $1\times1$ convolutional layers. The GRU encoder spatio-temporally encodes the last three frames of the incoming pose sequence, which are mapped to the size of the flattened heatmap by the MLP. After reshaping the output of the MLP to match that of the heatmap, we pass this to the convolution layers to get our raw heatmap. The final heatmap is obtained by capping this output with a sigmoid layer.
We use OpenTSNE's implementation of t-SNE \cite{openTSNE} which also implements the transform function, a feature missing in the original t-SNE variants. Finally, the codebook can be implemented as a tensor or as a dictionary, since the codebook serves as a lookup table where the queries (or keys) are locations on the heatmap of type integer. 

\section{Dataset: Details}
\label{app:dataset}
Human3.6M~\cite{ionescu2013human3} consists of motion-captured poses of seven publicly available subjects performing 15 actions. We follow the protocol proposed by \cite{barquero2023belfusion}. The first five publicly available subjects (S1, S5, S6, S7, S8) of the dataset are used for training, and the last two (S9, S11) for testing.  The dataset consists of 32 keypoints in total, from which 17 are selected. We zero-center them around the pelvis joint, and thus the the remaining 16  joints are forecasted with respect to the pelvis. Videos of this dataset have been recorded at 50 fps, and we take 0.5 seconds (25 frames) as input and forecast the next 2 seconds (100 frames). 
    
AMASS~\cite{mahmood2019amass} is a collection of various datasets containing 3D human poses. Following \cite{barquero2023belfusion}, we utilize 11 sets (406 subjects) from this collection for training and 7 sets (54 subjects) for testing.  The dataset contains videos at 60 fps after downsampling. We use 0.5 seconds (30 frames) as observation and forecast the next 2 seconds (120 frames). We also downsampled the input data of AMASS by increasing the stride to reduce the training time.

\section{Additional Quantitative Results}

\begin{figure*}
\centerline{\includegraphics[width=0.6\paperwidth]{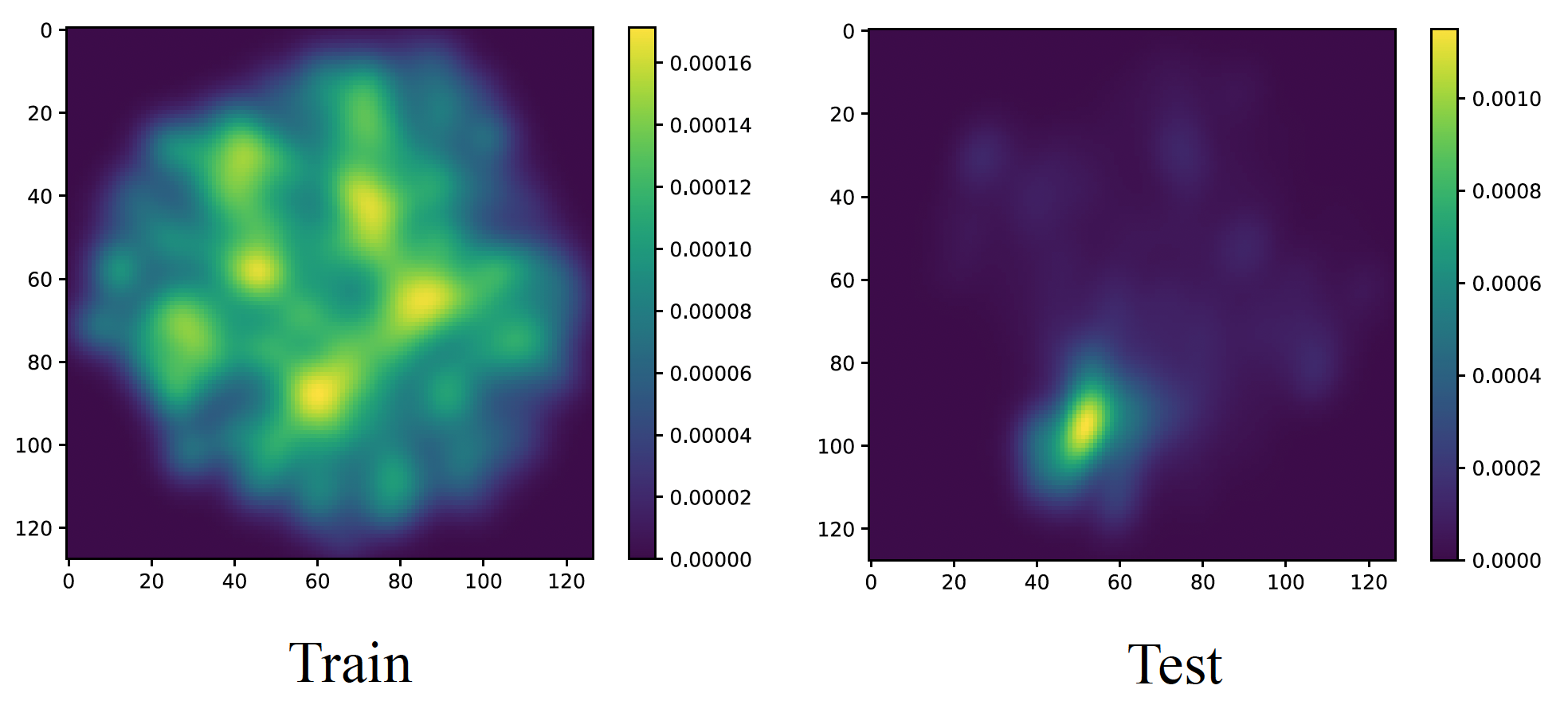}}
\caption{We plot the density map of ground truth sequences $Y$ for the training and testing split of AMASS suggested by \cite{barquero2023belfusion}. We observe that the splits can be highly imbalanced, and have a significant impact on determining the multimodal ground truth for a sample.}
\label{fig:hmyo2}
\end{figure*}

We report our results by restricting the multimodal ground truth to the testing split only. We observe that across both datasets the quantitative results are similar across different methods. While DivSamp is highly diverse, this does not necessarily translate to accurately predicting possible futures. A major observation is that while MotionMap is much more effective in recalling transitions from the test set (Table \ref{table:q2}), this does not come at the cost of general performance, as evident by these results (\ref{table:app_h36m}. \ref{table:app_amass}). Finally, we note that restricting the multimodal ground truth to the testing split limits the diversity of modes in the ground truth. In \Cref{fig:hmyo2}, we demonstrate that the AMASS testing dataset does not adequately represent the training data, with the testing multimodal ground truth missing the majority of modes. Assuming that the test split contains only five samples, each test sample would have between one and five multimodal ground truths. Furthermore, a discrepancy in the distributions of the train and test split means that the multimodal ground truths for the test set share no commonalities with the train set.

\begin{table*}[!t]
  \caption{Human3.6M dataset: All baselines are limited to 5 forecasts. Our method, unconstrained by the number of modes, is adjusted to produce an equal number of predictions. Metrics are reported in meters.}
  \label{table:app_h36m}
  \centering
\small
\begin{tabular}{lccccc}
\toprule
Method & Diversity ($\downarrow$) & ADE ($\downarrow$) & FDE ($\downarrow$) & MMADE ($\downarrow$) & MMFDE ($\downarrow$)  \\
\midrule
\rowcolor{Gray}Zero-Velocity & 0.000 & 0.597 & 0.884 & 0.616 & 0.884  \\ 
 TPK~\cite{walker2017theposeknows} & 6.727 & 0.568 & 0.757 & 0.582	& 0.756  \\ 

\rowcolor{Gray}DLow~\cite{yuan2020dlow} & 11.687 &  0.602 & 0.818 & 0.616 & 0.818   \\  

GSPS~\cite{mao2021gsps} & 14.729 & 0.584 & 0.791 & 0.602 & 0.791 \\ 

 \rowcolor{Gray}DivSamp~\cite{dang2022diverse} & 15.571 & 0.545 & 0.782	& 0.574 & 0.787  \\ 

BeLFusion~\cite{barquero2023belfusion} & 7.323 & 0.472 & 0.656 & 0.497 & 0.661 \\

 \rowcolor{Gray} CoMusion~\cite{sun2024towards} & 7.624 & 0.460 & 0.678 & 0.505 & 0.687 \\
\midrule
\textbf{MotionMap} & 8.190  & 0.491 & 0.642 & 0.505  & 0.643 \\
\bottomrule
\end{tabular}
\end{table*}

\begin{table*}[!t]
  \caption{AMASS dataset: All baselines are limited to 6 forecasts. Our method, unconstrained by the number of modes, is adjusted to produce an equal number of predictions. Metrics are reported in meters.}
  \label{table:app_amass}
  \centering
  \small
  \begin{tabular}{lccccc}
\toprule
Method&Diversity ($\downarrow$) & ADE ($\downarrow$) & FDE ($\downarrow$) & MMADE ($\downarrow$) & MMFDE ($\downarrow$) \\
\midrule
\rowcolor{Gray}Zero-Velocity & 0.000 & 0.755 & 0.992 & 0.776 & 0.998 \\ 
TPK~\cite{walker2017theposeknows} & 9.284 & 0.762 & 0.867 & 0.763 & 0.864  \\ 
\rowcolor{Gray}DLow~\cite{yuan2020dlow} & 13.192 & 0.739 & 0.842 & 0.733 & 0.846  \\ 
 GSPS~\cite{mao2021gsps} & 12.472 & 0.736 & 0.872 & 0.741 & 0.871 \\ 
\rowcolor{Gray}DivSamp~\cite{dang2022diverse} & 24.723	& 0.795 & 0.926 & 0.801 & 0.928 \\ 
 BeLFusion~\cite{barquero2023belfusion} & 9.643 & 0.620 & 0.751 & 0.632 & 0.751 \\ 
\rowcolor{Gray}CoMusion~\cite{sun2024towards} & 10.854 & 0.601 & 0.768 & 0.629 & 0.797 \\

\midrule

\textbf{MotionMap} & 9.483  & 0.624 & 0.729 & 0.643  & 0.736 \\

\bottomrule
\end{tabular}
\end{table*}

\section{Additional Qualitative Results}

We have provided some examples of generated future forecasts in the format of GIFs which are included in the supplementary materials in a folder called: \textbf{GIFs}. In the aforementioned visualization, the color blue refers to the input pose sequence, and red to the corresponding future. 

\subsection{Controllability}
\label{app:controllability}
\begin{figure*}
\centerline{\includegraphics[width=2.3\columnwidth]{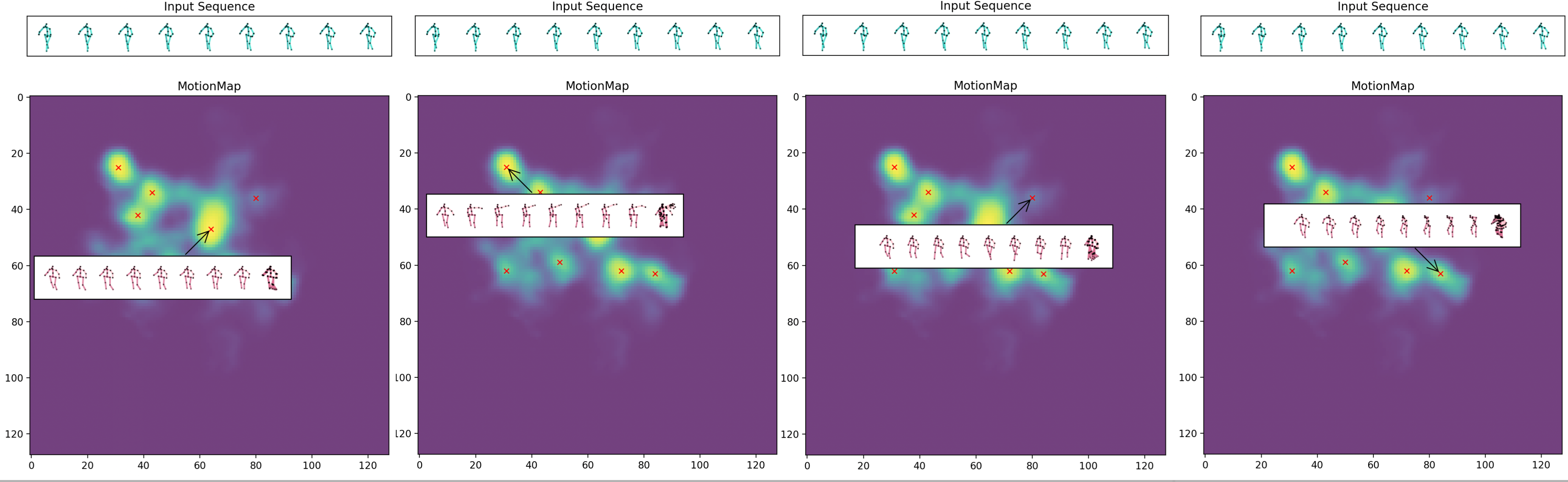}}
\caption{We visualize the modes (in red crosses) predicted by MotionMap. By hovering over the demo tool, we can view the decoded future poses corresponding to the given input pose sequence. We have uniformly selected eight frames in each sequence to demonstrate the motion and \textbf{stacked them on top of each other at the end} (the frame on the very right of each visualized sequence) to represent the amount of motion in each sequence.}
\label{fig:hover1}
\end{figure*}

\begin{figure*}
\centerline{\includegraphics[width=2.3\columnwidth]{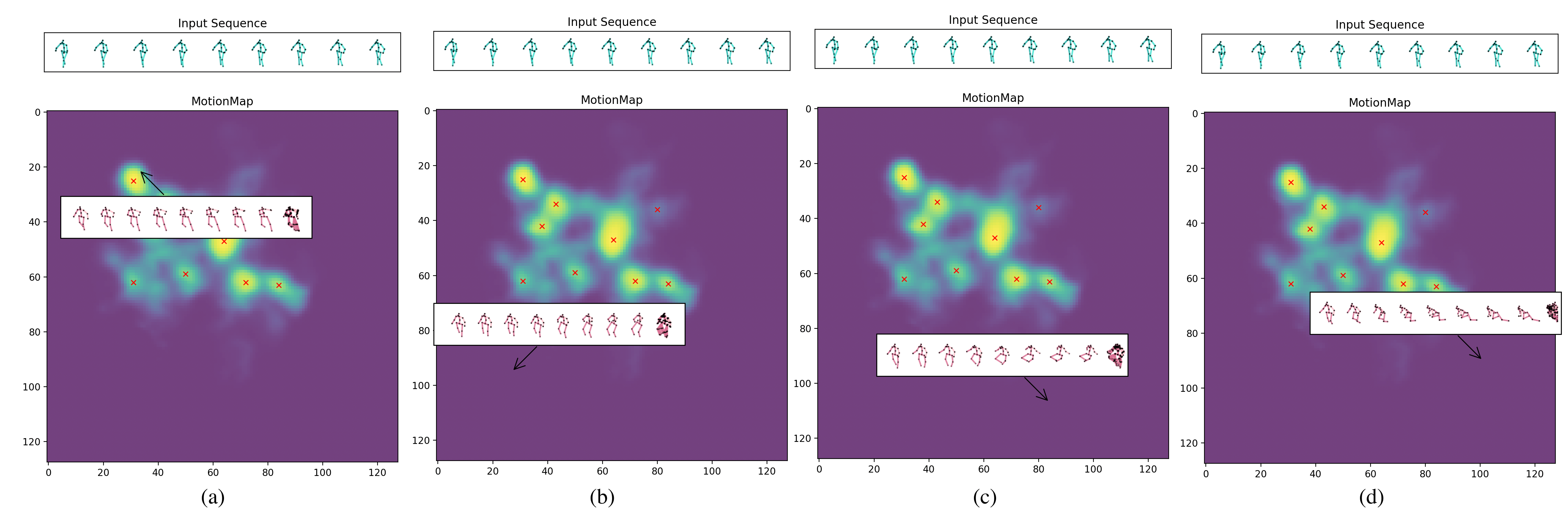}}
\caption{We show different strategies for controlled selection of \textit{non-maxima} forecasts: (a) Selecting samples in the vicinity of a model selected mode. (b,c,d) Based on the distribution of action labels. For instance, we could generate futures for rarer transitions such as sitting down (b) on a chair (c) on the floor, or (d) lying on the floor.}
\label{fig:hover2}
\end{figure*}

Our method enables control over the selection of modes. With the predicted MotionMap and its identified local maxima, we can focus solely on the most probable futures (\Cref{fig:hover1}) or, if needed, select a less likely future (using metadata) as required by the application's requirements (\Cref{fig:hover2}). To better show this possibility we have provided a demo. 

\subsection{Uncertainty}
\label{app:uncertainty}
\begin{figure*}
\centerline{\includegraphics[width=1.05\linewidth]{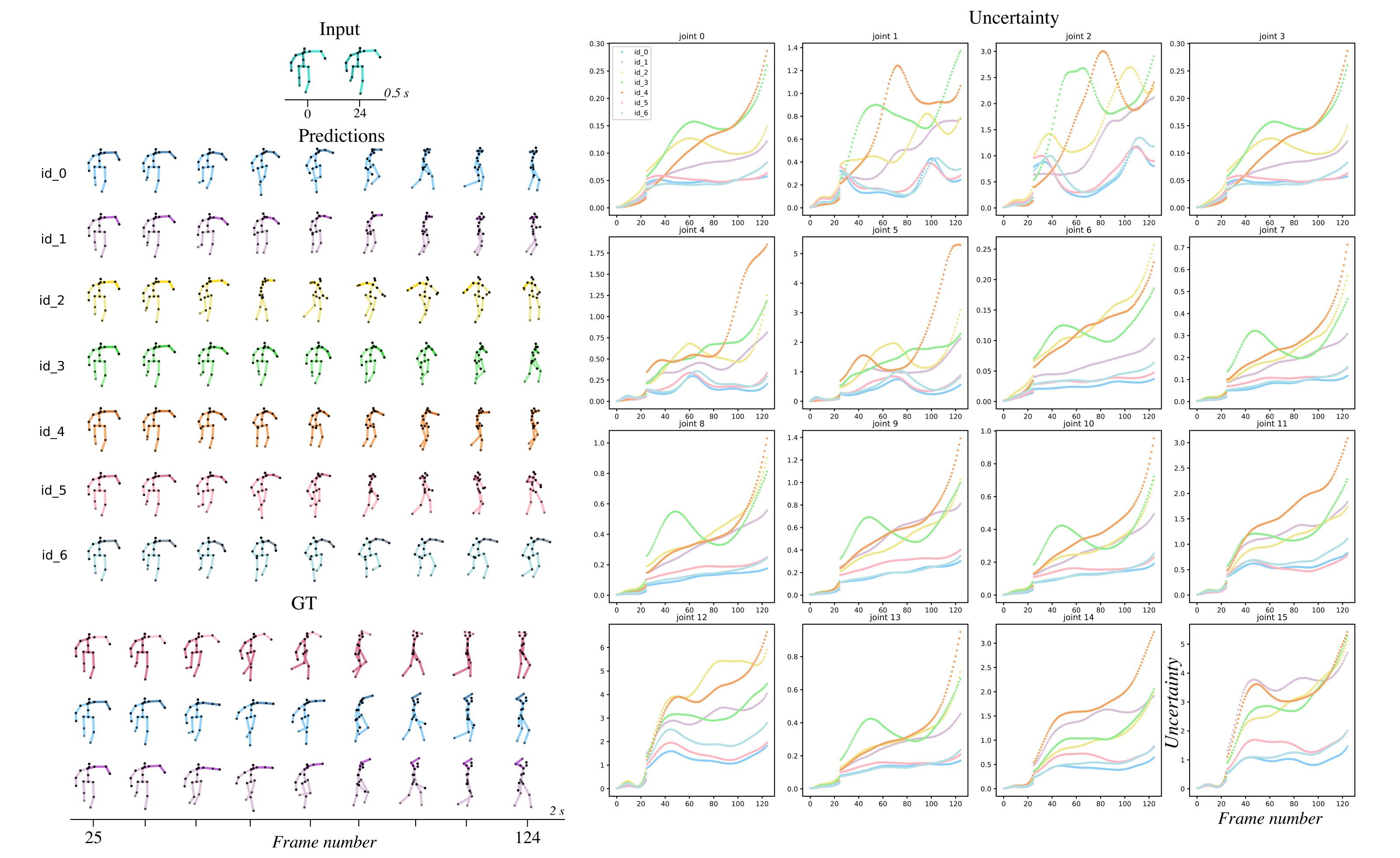}}
\caption{We show additional forecasts along with the predicted uncertainty per joint and time frame.}
\label{fig:uncer}
\end{figure*}

We have illustrated the predicted uncertainty plots for all the future predicted poses and the reconstructed past in \Cref{fig:uncer}. It is observable that the model is more certain about reconstructing the past since it is encoded as the input. The various trends in uncertainty demonstrate the dependency of the predicted uncertainty on the motion. Furthermore, joints that have greater movement or are further from the pelvis experience higher levels of uncertainty.

\subsection{Sampling Comparison}
\label{app:sampling}
\begin{figure*}
\vspace*{-2cm}
\centerline{\includegraphics[width=2.2\columnwidth]{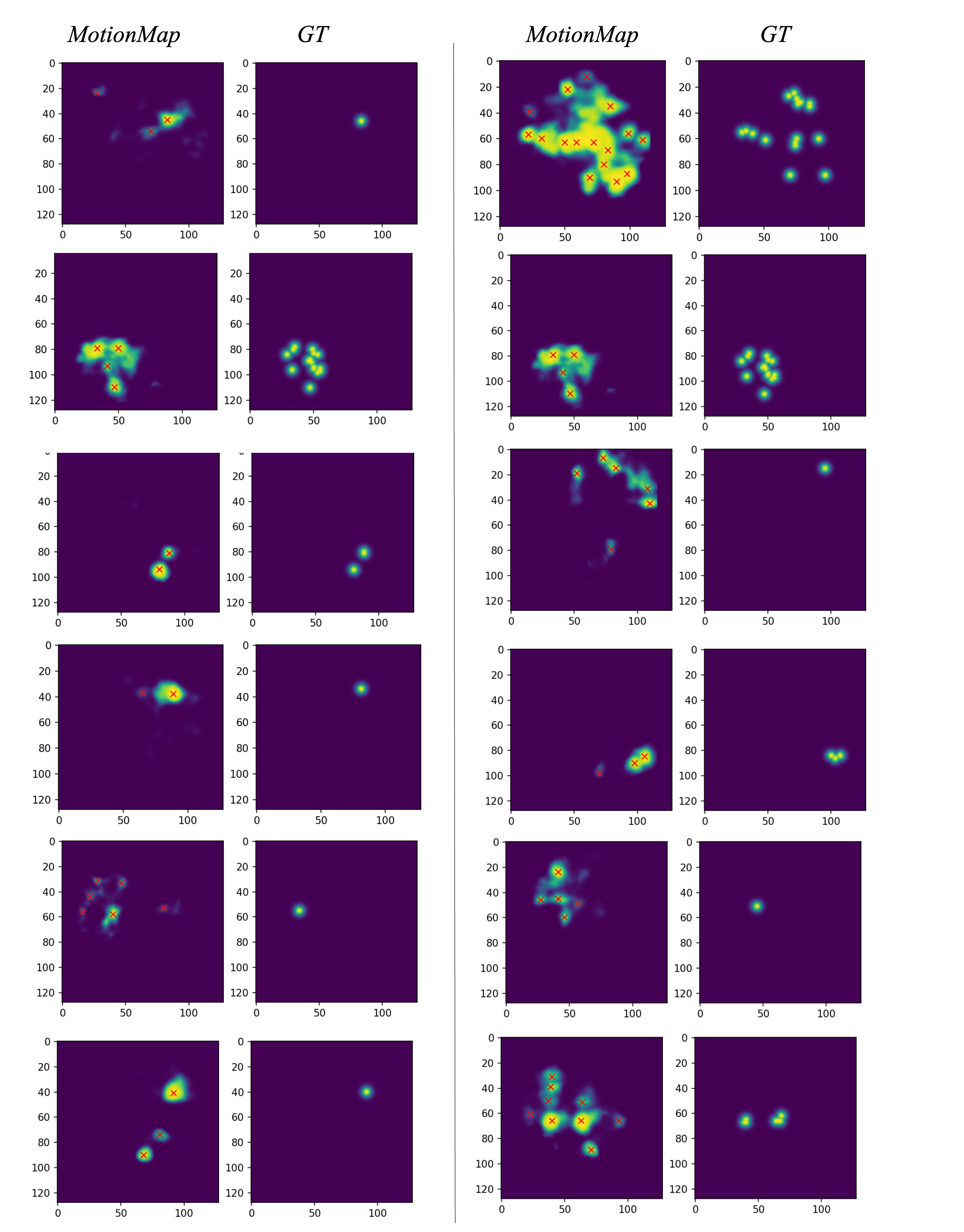}}
\caption{Qualitative comparison between MotionMap and the ground truth multimodal heatmap. Our observations indicate that MotionMap effectively captures the diversity of the modeled scenarios. The presence of a larger number of peaks in MotionMap is a result of learning generalized behaviour across the training split.}
\label{fig:hm_mm}
\end{figure*}

We compare the predicted MotionMaps with the ground truth heatmaps in \Cref{fig:hm_mm}. MotionMap is encouraged to predict a higher number of modes than present in the ground truth to identify rare transitions. Our visualizations confirms that MotionMap identifies other transitions while not missing out on the original ground truth motions. These miscellaneous transitions are learned by the MotionMap model from trends across the dataset.

How well can state-of-the-art baselines predict multimodality without explicitly encoding multimodal transitions? To study this, we collected predictions for each of the baselines for each input pose seqeuence. We then encode these pose forecasts into two dimensions as described in Section \ref{sec:dimred}. Next, we overlay them on the ground truth heatmap to identify the differences in the predictions and the ground truth. We observe that baselines that rely on anchors although diverse predict transitions which are unlikely for the given pose sequence. This also tallies with our quantitative evaluation. While this effect is reduced for diffusion based baselines, the methods are less diverse and do not capture rare modes. In contrast, MotionMap captures both common and rare mode since they are encoded in the form of local maxima.

\begin{figure*}
\vspace*{-2cm}
\centerline{\includegraphics[width=1.5\columnwidth]{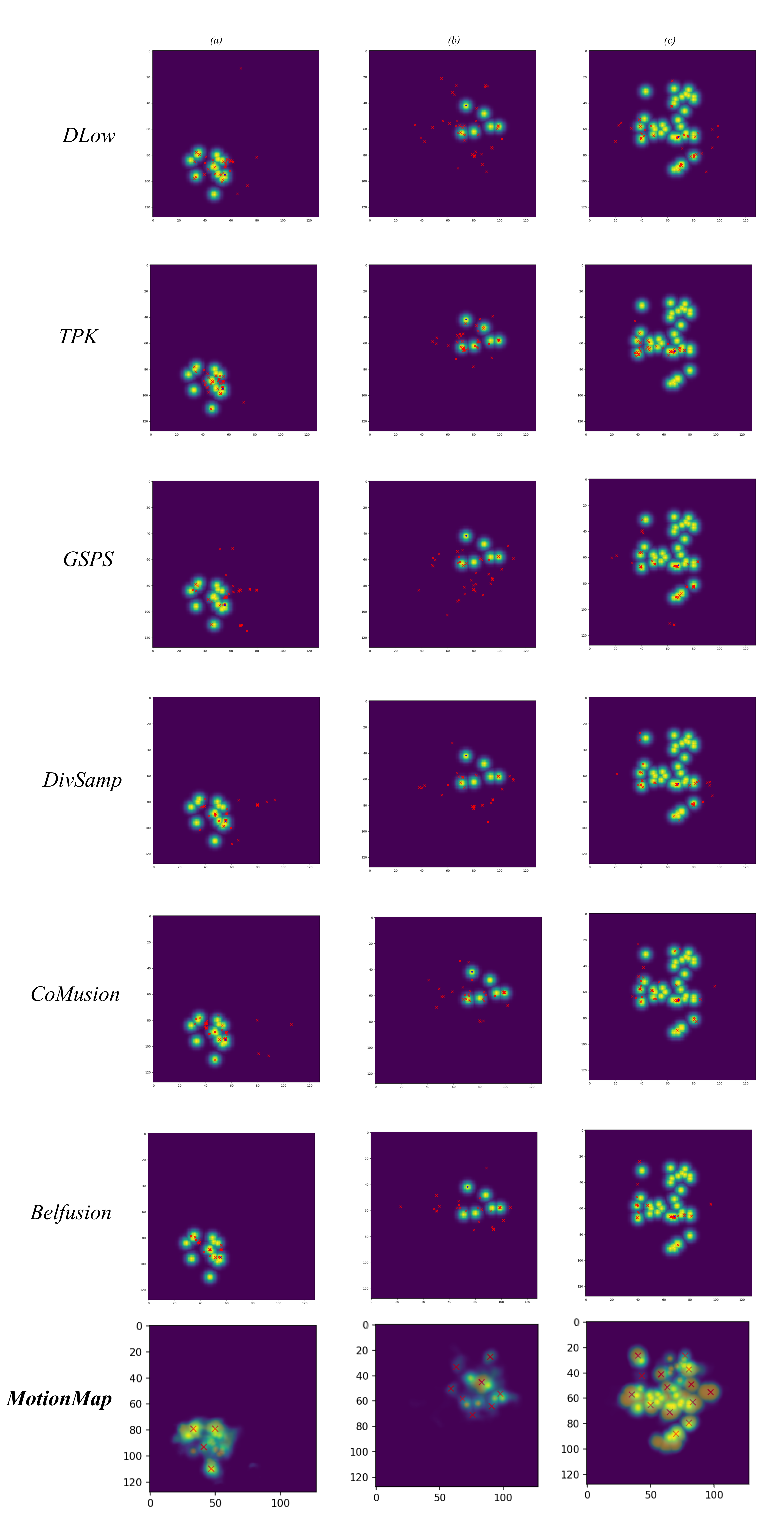}}
\caption{We overlay predictions for each baseline on the ground truth heatmap, for each of the three input pose sequences. The encoding of these predictions is shown as red crosses. For MotionMap, we directly overlay the predicted MotionMap (with crosses for maxima) on the ground truth heatmap. We note that methods are either highly diverse but unrealistic or are less diverse but predict likely futures. In contrast, MotionMap predicts both: common and rare modes since both are explicitly encoded in the training process.}
\label{fig:hm_all}
\end{figure*}

\end{document}